\documentclass[11pt]{article}

\usepackage[final]{acl}

\usepackage{times}
\usepackage{latexsym}
\usepackage{booktabs}
\usepackage{multirow}
\usepackage[T1]{fontenc}

\usepackage[utf8]{inputenc}

\usepackage{microtype}

\usepackage{inconsolata}

\usepackage{graphicx}

\usepackage{fontawesome5}
\usepackage{hyperref}

\usepackage{listings}
\usepackage{subcaption}
\usepackage{adjustbox}
\usepackage{xcolor}  

\definecolor{myblue}{HTML}{C9DAF8}
\definecolor{mypink}{HTML}{F4CCCC}  

\definecolor{questionbg}{HTML}{EEF4FB}
\definecolor{questionframe}{HTML}{4A90D9}
\definecolor{answerbg}{HTML}{FFF8EC}
\definecolor{answerframe}{HTML}{F5A623}
\definecolor{labelbg}{HTML}{F4F4F8}
\definecolor{labelframe}{HTML}{7B68EE}

\definecolor{badgeclarity}{HTML}{4A90D9}
\definecolor{badgeevasion}{HTML}{E74C3C}
\definecolor{badgepresident}{HTML}{8E44AD}
\definecolor{badgetrue}{HTML}{27AE60}
\definecolor{badgefalse}{HTML}{E74C3C}
\definecolor{badgeneutral}{HTML}{7F8C8D}

\usepackage{booktabs}
\usepackage{tabularx}
\usepackage{array}
\usepackage{multirow}
\usepackage{threeparttable}
\usepackage{makecell}
\usepackage{ragged2e}
\usepackage{placeins}
\usepackage{adjustbox}

\usepackage{pifont}

\usepackage{acro}

\usepackage[most]{tcolorbox}
\usepackage{todonotes}

\usepackage{tcolorbox}
\usepackage{xcolor}
\usepackage{soul}
\usepackage{fontawesome5} 
\usepackage{tikz}

\usepackage{seqsplit}

\tcbuselibrary{skins, breakable}

\newcommand{\badge}[3]{%
  \tikz[baseline=-0.5ex]{
    \node[
      fill=#1!15,
      draw=#1,
      rounded corners=3pt,
      inner sep=3pt,
      font=\tiny\sffamily\bfseries
    ] {%
      \textcolor{#1!70!black}{#2:}~\textcolor{black}{#3}%
    };
  }%
}

\tcbset{
  qabox/.style={
    enhanced,
    boxrule=1.5pt,
    arc=6pt,
    left=8pt, right=8pt, top=6pt, bottom=6pt,
    fonttitle=\small\bfseries\sffamily,
    attach boxed title to top left={yshift=-2mm, xshift=8pt},
    boxed title style={enhanced, arc=4pt, boxrule=0pt, interior style={fill=white}},
    drop shadow={gray!25!white},
  }
}

\tcbset{colback=yellow!10!white, colframe=orange!50!black, boxrule=0.5pt, arc=3pt, left=6pt, right=6pt}

\lstdefinestyle{aclprompt}{
    basicstyle=\ttfamily\tiny,
    moredelim=[is][\boldred]{**}{**},   
backgroundcolor=\color{questionbg},      
    rulecolor=\color{questionframe},        %
    frame=single,
    framesep=1.2mm,
    framerule=2pt,
    breaklines=true,
    breakatwhitespace=true,
    columns=flexible,
    keepspaces=true,
    showstringspaces=false,
    inputencoding=utf8,
    extendedchars=true,
    captionpos=b,
    xleftmargin=0.05cm,
    xrightmargin=0.05cm,
    aboveskip=0.2em,
    belowskip=0.2em,
    linewidth=\linewidth,
    literate=
        {à}{{`a}}1 {è}{{\`e}}1 {ì}{{`i}}1 {ò}{{\`o}}1 {ù}{{\`u}}1
        {À}{{`A}}1 {È}{{\`E}}1 {Ì}{{`I}}1 {Ò}{{`O}}1 {Ù}{{\`U}}1
        {á}{{\'a}}1 {é}{{\'e}}1 {í}{{\'i}}1 {ó}{{\'o}}1 {ú}{{\'u}}1
        {Á}{{\'A}}1 {É}{{\'E}}1 {Í}{{\'I}}1 {Ó}{{\'O}}1 {Ú}{{\'U}}1
}

\newcommand{\boldred}[1]{\textcolor{red}{\textbf{#1}}}

\newcommand{\sd}[1]{\ensuremath{\text{\scriptsize(#1)}}}

\DeclareAcronym{cd}{
  short = CD ,
  long  = cognitive distortion ,
  short-plural = s ,
}
\DeclareAcronym{cbt}{
  short = CBT ,
  long  = cognitive behavioural therapy 
}
\DeclareAcronym{llm}{
  short = LLM ,
  long  = large language model 
}
\DeclareAcronym{nlp}{
  short = NLP ,
  long  = natural language processing 
}
\DeclareAcronym{iaa}{
  short = IAA ,
  long  = inter-annotator agreement
}
\DeclareAcronym{ce}{
  short = CE ,
  long  = cross-entropy
}

\title{KCLarity at SemEval-2026 Task 6:\\ Encoder and Zero-Shot Approaches to Political Evasion Detection}

\author{Archie Sage\thanks{Equal contribution.} \\
  King's College London \\
  \texttt{archie.sage@kcl.ac.uk} \\\And
  Salvatore Greco\footnotemark[1] \\
  King's College London \\
  \texttt{salvatore.greco@kcl.ac.uk} \\
  }

\begin{document}
\maketitle

%
%

\begin{abstract}

This paper describes the KCLarity team's participation in CLARITY, a shared task at SemEval 2026 on classifying ambiguity and evasion techniques in political discourse. We investigate two modelling formulations: (i) directly predicting the clarity label, and (ii) predicting the evasion label and deriving clarity through the task taxonomy hierarchy. We further explore several auxiliary training variants and evaluate decoder-only models in a zero-shot setting under the evasion-first formulation. Overall, the two formulations yield comparable performance. Among encoder-based models, RoBERTa-large achieves the strongest results on the public test set, while zero-shot GPT-5.2 generalises better on the hidden evaluation set.

\end{abstract}

%
%

\section{Introduction}

Public scrutiny of politicians depends not only on access to questioning but also on the clarity of their responses. Prior work shows that politicians exhibit significantly lower clear reply rates than non-politicians during televised interviews \citep{bull2003microanalysis}. Such unclear responses are commonly described as \textit{equivocation} or \textit{evasion} in the political communication literature \citep{watzlawick2011pragmatics, bavelas1988political}. These findings motivate the development of effective 
and low-cost automated methods for identifying unclear or evasive answers.

The CLARITY \citep{thomas2026semeval2026task6clarity} shared task at SemEval 2026 focuses on developing \ac{nlp} methods for detecting and classifying response ambiguity and evasion strategies in political discourse. The shared task consists of two subtasks: predicting response clarity (Task 1) and identifying evasion strategies (Task 2).

In this paper, we present the systems developed by the KCLarity team for the CLARITY shared task. We evaluate fine-tuned encoder-based models and decoder-only models prompted in a zero-shot setting. For the encoder models, we consider two training targets: (i) predicting clarity labels directly (\textit{direct clarity}) and (ii) predicting evasion labels and inferring clarity via the hierarchical taxonomy (\textit{evasion-based clarity}). We further explore training configurations including per-class loss weighting (Section~\ref{sec:loss_weighting}), data-splitting strategies (Section~\ref{sec:data_splits}), input representations (Section~\ref{subsec:input-representation}), person-name masking (Appendix~\ref{sec:masking_appendix}), and additional exploratory experiments (Appendix~\ref{sec:exploratory}). For the decoder models, we evaluate zero-shot prediction with both open-weight and commercial models.

On the public test split, our strongest encoder-based model  - RoBERTa-large trained to predict evasion labels and mapped to clarity via the taxonomy - performs competitively on both tasks and outperforms the decoder-only models in our zero-shot evaluation. Among zero-shot systems, GPT-5.2 is the strongest, and we use the same evasion-first formulation for consistency. In the official shared task evaluation on the hidden test set, our top submission is the zero-shot GPT-5.2 system, ranking 22nd out of 44 in Task 1 (macro F1 = 0.74) and 13th out of 33 in Task 2 (macro F1 = 0.50).\footnote{Our implementation code is available at: \url{https://github.com/semeval-2026-kclarity/clarity}}

%
%

\section{Background}
\label{sec:background}

%
%

\subsection{Task Definition and Dataset}

\begin{table}[]
\centering
\footnotesize
\caption{Taxonomy of response clarity classification.}
\label{tab:taxonomy}
\begin{tabularx}{\columnwidth}{lX}
\toprule
\textbf{Clarity Level} & \textbf{Evasion Level} \\
\midrule
Clear Reply      & Explicit \\
Ambivalent Reply & Implicit, Dodging, General, Deflection, Partial \\
Clear Non-Reply  & Declining, Ignorance, Clarification \\
\bottomrule
\end{tabularx}
\end{table}

The CLARITY task comprises two subtasks in English political discourse: predicting the response's clarity level (Task 1) and identifying the evasion technique (Task 2). The tasks are hierarchically related through the taxonomy introduced in the QEvasion dataset \citep{thomas2024isaidthatdataset}. The mapping between the two label sets is shown in Table~\ref{tab:taxonomy}.

The QEvasion dataset comprises 3,448 training instances and 308 test instances, drawn from US presidential interviews.\footnote{An example question–answer pair is shown in Appendix~\ref{apx:data-analysis}.} 
Each instance is annotated with a single clarity label. For the evasion level, training instances receive one evasion label, while test instances contain three evasion labels assigned independently by separate annotators. This multi-annotator supervision in the test set informs the evaluation strategy described in Section~\ref{sec:eval_metrics}. In addition to these primary labels, each instance includes further metadata, such as the interview date, the president, whether the question contains multiple sub-questions, and whether the question is affirmative.

%
%

\subsection{Class Imbalance and Inter-annotator Agreement}
\label{sec:class_imbalance}

As noted by the shared task organisers, clarity-level \ac{iaa}, measured using Fleiss’ Kappa $\kappa$ \citep{fleiss1971measuring} on the QEvasion dataset shows that annotators (i) almost never confuse \textit{Clear Reply} with \textit{Clear Non-Reply} ($\kappa = 0.97$), (ii) sometimes disagree between \textit{Clear Reply} and \textit{Ambivalent} ($\kappa = 0.65$), and (iii) also sometimes disagree between \textit{Clear Non-Reply} and \textit{Ambivalent} ($\kappa = 0.71$) \citep{thomas2024isaidthatdataset}. This indicates that the main modelling challenge at clarity-level classification lies in the overrepresented \textit{Ambivalent} class, which accounts for 59.2\% of samples, compared with \textit{Clear Reply} and \textit{Clear Non-Reply} at 30.5\% and 10.3\% prevalence. 
Accordingly, we explore loss weighting to mitigate the resulting class imbalance (Section~\ref{sec:loss_weighting}) amongst other strategies.

%
%

\section{System Overview}
\label{sec:system_overview}

We evaluate two approaches for the CLARITY task: (i) fine-tuning encoder-based models and (ii) prompting decoder-only models in a zero-shot setting (Section~\ref{subsec:models}). For the encoder models, we explore loss-weighting strategies (Section~\ref{sec:loss_weighting}), alternative input representations (Section~\ref{subsec:input-representation}), and two prediction targets: predicting clarity and evasion separately, or predicting evasion and inferring clarity via the label hierarchy (Section~\ref{subsec:direct-vs-evasion}).

%
%

\subsection{Models}
\label{subsec:models}

\paragraph{Encoder-based models.} We fine-tuned RoBERTa \citep{zhuang-etal-2021-robustly} and DeBERTa-v3 \citep{he2023debertav3improvingdebertausing}, evaluating base and large variants. We also conducted preliminary single-seed experiments with BERT \citep{devlin-etal-2019-bert}, ELECTRA \citep{Clark2020ELECTRA}, and other encoder models, but report results only for RoBERTa and DeBERTa-v3, as the remaining models were consistently less competitive.

\paragraph{Decoder-based models.} We prompted in zero-shot settings \citep{dong-etal-2024-survey} decoder-only models from different families and sizes, including open-weight models such as Llama 3~\citep{grattafiori2024llama3herdmodels,llama3modelcard}, Qwen~\citep{qwen3technicalreport}, Gemma 3~\citep{gemma_2025}, and a commercial model GPT-5.2\footnote{Snapshot: \texttt{gpt-5.2-2025-12-11}} \citep{singh2025openaigpt5card}.

%
%

\subsection{Loss Weighting}
\label{sec:loss_weighting}

To mitigate class imbalance in QEvasion (Section~\ref{sec:class_imbalance}), we train encoder models with weighted \ac{ce}, defined as follows:
\begin{equation*}
\mathcal{L}_{\mathrm{WCE}}(x,y)
= -\, w_y \log p(y \mid x)
\end{equation*}
Additionally, we compare three weighting schemes. (i) \textit{Unweighted}: $w_y = 1$ for all classes. 
(ii) \textit{Balanced}: the standard inverse-frequency weight 
$w_y = \frac{N}{C \cdot n_y}$, where $n_y$ is the class count, 
$N$ the total number of training instances, and $C$ the number of classes. 
(iii) \textit{Sqrt}: a milder reweighting 
$w_y = \frac{1}{\sqrt{f_y + \epsilon}}$, where $f_y = n_y/N$ is the class frequency and $\epsilon$ a small stability constant. 
For the sqrt scheme, weights are capped and rescaled to unit mean to prevent extreme upweighting of rare classes. 
All weights are computed from and applied only to the training split.

%
%

\subsection{Input Representation}
\label{subsec:input-representation}

Each instance consists of a question-answer pair. We compare two encoder input formats that differ in both field ordering and boundary representation.

\paragraph{Segmented representation (answer first).}
The answer and question are encoded as two segments, with the answer provided first:
\[
\texttt{[CLS] \; a \; [SEP] \; q \; [SEP]}.
\]
When supported, token-type embeddings indicate the segment boundary.

\paragraph{Marked representation (question first).}
Alternatively, the question and answer are concatenated into a single sequence with explicit marker tokens:
\[
\texttt{[QUESTION] \; q \; [ANSWER] \; a}.
\]
Here, the boundary between the two texts is indicated by learned special tokens. We add \texttt{[QUESTION]} and \texttt{[ANSWER]} as special tokens and resize the model's input embeddings accordingly.

%
%

\subsection{Direct vs Evasion-Based Clarity}
\label{subsec:direct-vs-evasion}

\begin{table}[t]
  \centering
  \caption{\textbf{RoBERTa models: direct vs evasion-based clarity on the public test split.}
Macro-F1 (F1), precision (P), and recall (R), averaged over three seeds (standard deviation in brackets). 
Higher scores between the two formulations are shown in bold.}
  \label{tab:roberta_direct_vs_evasion_based_clarity}

  \footnotesize
  \setlength{\tabcolsep}{3.5pt}
  \renewcommand{\arraystretch}{1.08}

  \begin{tabular*}{\columnwidth}{@{\extracolsep{\fill}} l ccc ccc @{}}
    \toprule
    \multirow{2}{*}{\textbf{Model}} &
    \multicolumn{3}{c}{\textit{Direct clarity}} &
    \multicolumn{3}{c}{\textit{Evasion-based clarity}} \\
    \cmidrule(lr){2-4} \cmidrule(lr){5-7}
    & \textbf{F1} & \textbf{P} & \textbf{R}
    & \textbf{F1} & \textbf{P} & \textbf{R} \\
    \midrule

    \shortstack[l]{base}
      & \textbf{0.598} & 0.586 & \textbf{0.636}
      & 0.595 & \textbf{0.589} & 0.619 \\[-0.2ex]
      & \sd{0.010} & \sd{0.013} & \sd{0.021}
      & \sd{0.028} & \sd{0.035} & \sd{0.030} \\[0.55ex]

    \shortstack[l]{large}
      & 0.658 & 0.681 & \textbf{0.661}
      & \textbf{0.661} & \textbf{0.694} & 0.641 \\[-0.2ex]
      & \sd{0.024} & \sd{0.081} & \sd{0.030}
      & \sd{0.022} & \sd{0.013} & \sd{0.023} \\

    \bottomrule
  \end{tabular*}
\end{table}

In line with the terminology in the dataset paper \citep{thomas2024isaidthatdataset}, we evaluate clarity either \textit{directly} (predicting the three clarity labels) or \textit{via evasion} (predicting the nine-way evasion labels and then mapping predictions to clarity using the taxonomy). Motivated by prior findings that evasion-based clarity can outperform direct clarity \citep{thomas2024isaidthatdataset}, we include an ablation on comparable RoBERTa-base and RoBERTa-large models.

Table~\ref{tab:roberta_direct_vs_evasion_based_clarity} shows mixed evidence on stability: direct clarity exhibits lower variance for the base model, while for the large model evasion-based clarity is comparably or more stable, particularly in precision. Performance differences are small: direct clarity tends to yield slightly higher recall, while evasion-based clarity yields slightly higher precision; macro F1 is within variance for both settings. Given these limited differences, we focus the remainder of our analysis on evasion-based clarity because (i) it performs comparably to direct clarity, and (ii) it enables a single model trained for evasion to be reused to obtain clarity labels via the mapping, eliminating the need to train a separate model for clarity prediction.

%
%

\section{Experimental Setup}
\label{sec:experimental-settings}

%
%

\subsection{Data Splits}
\label{sec:data_splits}

We evaluate models under two splitting regimes. In the \textit{label-stratified} setting, we preserve class distributions across training and validation; 
for Task~2, we apply \textit{dual stratification} over both evasion and mapped clarity labels. In the \textit{president-disjoint} setting, all responses from a given president appear exclusively in one split, preventing speaker leakage and testing cross-speaker generalisation.

Unless otherwise stated, all encoder-based models use an 80/20 dual-stratified split of the 3,448 training samples (2,758 training / 690 validation) and are evaluated on the 308 publicly available test instances. Decoder-only models are evaluated zero-shot on the public test set, so no train–validation split is applied to them. Official rankings are based on the shared task's hidden test set, with predictions submitted via CodaBench\footnote{\url{https://www.codabench.org/}} 
to ensure fully blind evaluation. An ablation comparing the two splitting strategies is reported in Appendix~\ref{app:strat_vs_president_disjoint}.

%
%

\subsection{Evaluation Metrics}
\label{sec:eval_metrics}

For clarity level classification (Task~1), each test instance carries a single gold label; we report macro-averaged F1 (F1), Precision (P), and Recall (R).

For evasion level classification (Task~2), each test instance is independently labelled by three annotators. Rather than collapsing this multi-annotator supervision via majority vote - an approach that can discard valuable signal from legitimate disagreement \citep{fleisig-etal-2023-majority, basile-etal-2021-need} - we retain all three annotations. We compute macro-F1 separately against each annotator (F1\textsubscript{A1}, F1\textsubscript{A2}, and F1\textsubscript{A3}) and their average (F1\textsubscript{avg}). We additionally report $\mathrm{ACC}_{\mathrm{match}}$, the fraction of predictions matching at least one annotator's label, capturing the assumption that each annotation constitutes a plausible interpretation.

%
%

\section{Results}
\label{sec:results}

In this section, we first present development-phase results for the fine-tuned encoder models (Section~\ref{sec:results-encoder}) and zero-shot decoder models (Section~\ref{subsec:results-zero-shot}). We then discuss the official system rankings (Section~\ref{subsec:results-official-ranking}) and error analysis (Section~\ref{subsec:results-error-analysis}).

%
%

\begin{table*}[t]
  \centering
  \caption{\textbf{Fine-tuned encoder results on the public test set.}
    For evasion-based clarity we report macro-F1 (F1), precision (P), and recall (R).
    For evasion we report $\mathrm{ACC}_{\mathrm{match}}$, the fraction of predictions matching at least one annotator,
    per-annotator macro-F1, and the average macro-F1 across annotators.
    Results are averaged over three seeds (mean on the main row, standard deviation in brackets on the row beneath).
    Best-performing metrics are shown in bold.}
  \label{tab:encoder_results}

  \small
  \begin{tabular*}{\textwidth}{@{\extracolsep{\fill}} l ccc c cccc}
    \toprule
    \multirow{2}{*}{\textbf{Model}} & \multicolumn{3}{c}{\textit{Evasion-based clarity}} & \multicolumn{5}{c}{\textit{Evasion}} \\
    \cmidrule(lr){2-4} \cmidrule(lr){5-9}
    & \textbf{F1} & \textbf{P} & \textbf{R} & $\mathbf{ACC}_{\mathbf{match}}$ & \textbf{F1\textsubscript{A1}} & \textbf{F1\textsubscript{A2}} & \textbf{F1\textsubscript{A3}} & \textbf{F1\textsubscript{avg}} \\
    \midrule

    RoBERTa-base
      & 0.595 & 0.589 & 0.619 & 0.516 & 0.349 & 0.332 & 0.333 & 0.338 \\[-0.2ex]
      & \sd{0.028} & \sd{0.035} & \sd{0.030} & \sd{0.014} & \sd{0.024} & \sd{0.007} & \sd{0.008} & \sd{0.011} \\[0.65ex]

    RoBERTa-large
      & \textbf{0.661} & \textbf{0.694} & \textbf{0.641} & 0.539 & \textbf{0.363} & \textbf{0.378} & \textbf{0.374} & \textbf{0.371} \\[-0.2ex]
      & \sd{0.022} & \sd{0.013} & \sd{0.023} & \sd{0.011} & \sd{0.030} & \sd{0.035} & \sd{0.047} & \sd{0.037} \\[0.65ex]

    DeBERTa-v3-base
      & 0.535 & 0.564 & 0.565 & 0.505 & 0.304 & 0.337 & 0.303 & 0.314 \\[-0.2ex]
      & \sd{0.067} & \sd{0.033} & \sd{0.030} & \sd{0.027} & \sd{0.014} & \sd{0.016} & \sd{0.015} & \sd{0.011} \\[0.65ex]

    DeBERTa-v3-large
      & 0.616 & 0.634 & 0.610 & \textbf{0.541} & 0.321 & 0.305 & 0.325 & 0.317 \\[-0.2ex]
      & \sd{0.018} & \sd{0.021} & \sd{0.031} & \sd{0.007} & \sd{0.006} & \sd{0.010} & \sd{0.006} & \sd{0.006} \\

    \bottomrule
  \end{tabular*}
\end{table*}

\begin{table*}[t]
  \centering
  \caption{\textbf{Zero-shot results on the public test set.}
    For evasion-based clarity we report macro-F1 (F1), precision (P), and recall (R).
    For evasion we report $\mathrm{ACC}_{\mathrm{match}}$, the fraction of predictions matching at least one annotator,
    per-annotator macro-F1, and the average macro-F1 across annotators.
    Best-performing metrics are shown in bold.}

  \label{tab:zero-shot-results}
  \resizebox{\textwidth}{!}{%
      \tiny
  \begin{tabular}{l ccc c cccc}
    \toprule
    \multirow{2}{*}{\textbf{Model}} & \multicolumn{3}{c}{\textit{Evasion-based clarity}} & \multicolumn{5}{c}{\textit{Evasion}} \\
    \cmidrule(lr){2-4} \cmidrule(lr){5-9}
    & \textbf{F1} & \textbf{P} & \textbf{R} & $\mathbf{ACC}_{\mathbf{match}}$ & \textbf{F1\textsubscript{A1}} & \textbf{F1\textsubscript{A2}} & \textbf{F1\textsubscript{A3}} & \textbf{F1\textsubscript{avg}} \\
    \midrule
    gemma\_3\_27b\_it & 0.413 & 0.501 & 0.446 & 0.344 & 0.137 & 0.149 & 0.137 & 0.141 \\
    gpt\_oss\_120b & 0.354 & 0.374 & 0.354 & 0.357 & 0.104 & 0.121 & 0.115 & 0.114 \\
    Llama-3.1-8B-Instruct & 0.346 & 0.385 & 0.370 & 0.351 & 0.054 & 0.079 & 0.068 & 0.067 \\
    Llama-3.3-70B-Instruct & 0.532 & 0.544 & 0.547 & 0.416 & 0.284 & 0.292 & 0.294 & 0.290 \\
        Qwen3-8B & 0.337 & 0.361 & 0.350 & 0.292 & 0.060 & 0.078 & 0.070 & 0.069 \\
    Qwen3-32B & 0.338 & 0.383 & 0.347 & 0.331 & 0.063 & 0.098 & 0.089 & 0.083 \\
    GPT-5.2 & \textbf{0.626} & \textbf{0.600} & \textbf{0.670} & \textbf{0.481} & \textbf{0.363} & \textbf{0.339} & \textbf{0.371} & \textbf{0.358} \\
    \bottomrule
  \end{tabular}}
  \end{table*}

 \begin{figure*}
    \centering
    \begin{subfigure}[t]{0.49\textwidth}
        \centering
        \caption{\hspace*{35pt}\textbf{(\alph{subfigure})\ Evasion-based clarity}}
        \vspace{2pt}
        \includegraphics[width=\linewidth]{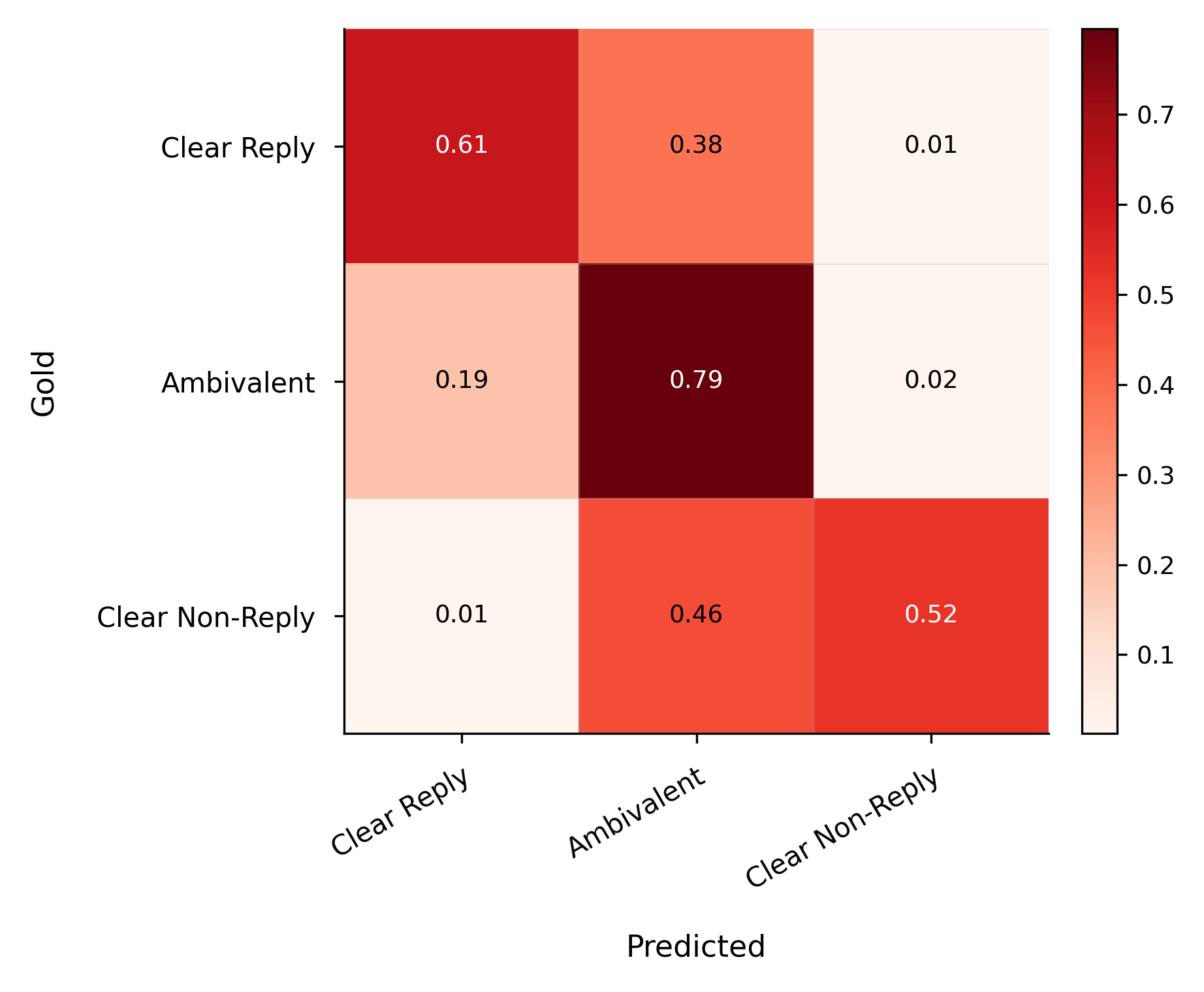}
        \label{fig:confusion_matrix_clarity}
    \end{subfigure}
    \hfill
    \begin{subfigure}[t]{0.49\textwidth}
        \centering
        \caption{\hspace*{30pt}\textbf{(\alph{subfigure})\ Evasion}}
        \includegraphics[width=\linewidth]{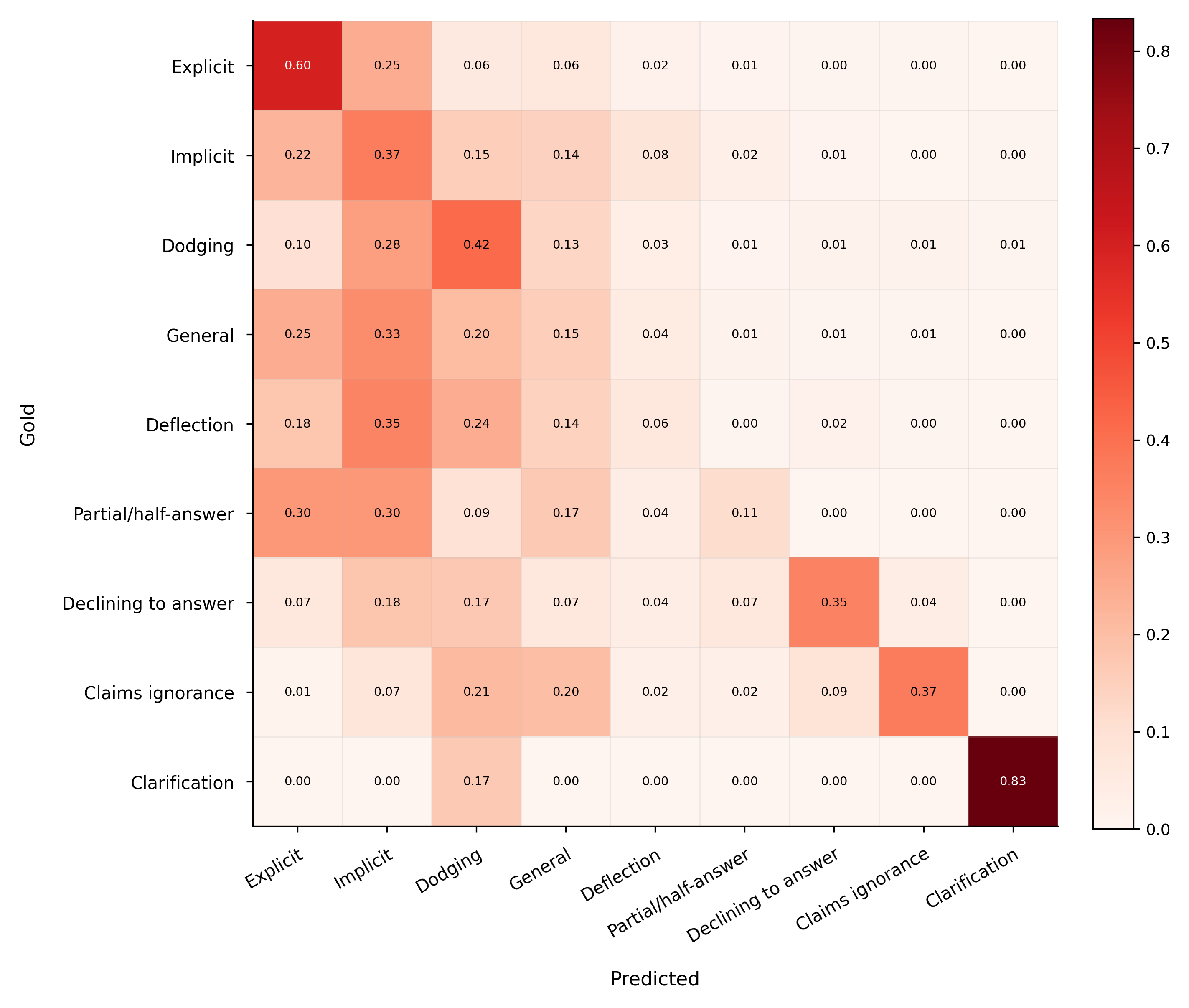}
        \label{fig:confusion_matrix_evasion}
    \end{subfigure}
    \vspace{-6mm}
    \caption[Confusion matrices]{\textbf{Row-normalised confusion matrices on the test set for RoBERTa-large model.} Rows correspond to gold labels and columns to predicted labels; values represent row-wise proportions such that each row sums to 1. Raw counts are averaged across random seeds and, for the evasion task, across annotators prior to row normalisation.}
    \label{fig:confusion_matrix}
\end{figure*}

\subsection{Fine-Tuned Encoder Results}
\label{sec:results-encoder}

Table~\ref{tab:encoder_results} reports results for RoBERTa and DeBERTa-v3 (base and large), which were fine-tuned to predict evasion labels with clarity inferred via the taxonomy. All models achieve moderate performance, with clarity macro F1 ranging from 0.535 to 0.661 and average evasion F1 from 0.314 to 0.371. RoBERTa-large is the best-performing model on both tasks (clarity F1 = 0.661; evasion F1\textsubscript{avg} = 0.371), approaching the fine-tuned LLaMA-70B baseline (F1 = 0.68) reported by \citet{thomas2024isaidthatdataset}. DeBERTa-v3-large is the second-best model and more stable across seeds; it achieves the highest $\text{ACC}_{\text{match}}$ (0.541), though this falls within RoBERTa-large's variance. Encoder models generally outperformed most zero-shot decoder models during development (Section~\ref{subsec:results-zero-shot}).

Several auxiliary strategies were explored but yielded no substantial or consistent gains: person-name masking (Appendix~\ref{sec:masking_appendix}), loss weighting via inverse-frequency and sqrt-based schemes (Appendix~\ref{app:loss_weighting}), intermediate fine-tuning on Earnings Calls Q\&A \citep{nuaimi-etal-2025-detecting} for cross-domain transfer (Appendix~\ref{app:exploratory_cross_domain}), and augmenting inputs with \ac{cd} probability buckets (Appendix~\ref{app:cd}). The latter two approaches additionally introduced training instability.


%
%

\subsection{Zero-Shot Results}
\label{subsec:results-zero-shot}

Table~\ref{tab:zero-shot-results} reports zero-shot results for the decoder models, all prompted to predict the evasion level with clarity inferred via the taxonomy. The same prompt was used for all models (Appendix~\ref{app:llm_prompt}). Among open-weight models, Llama-3.3-70B-Instruct performs best (clarity F1 = 0.532; evasion F1\textsubscript{avg} = 0.290), while smaller models across all families cluster well below. The commercial model GPT-5.2 substantially outperforms all open-weight alternatives on both tasks, achieving a clarity F1 of 0.626 and an evasion F1\textsubscript{avg} of 0.358 with an $\text{ACC}_{\text{match}}$ of 0.481. Nevertheless, all zero-shot decoder models underperform the best encoder models (Section~\ref{sec:results-encoder}), with GPT-5.2 being the only model approaching comparable performance.


%
%

\subsection{Official Ranking}
\label{subsec:results-official-ranking}

Official rankings were determined on the hidden evaluation set via CodaBench, as described in Section~\ref{sec:data_splits}. We submitted two systems for both tasks: (1) a majority-vote ensemble of five RoBERTa-large models trained on evasion labels, with clarity inferred via the taxonomy, and (2) zero-shot GPT-5.2 with the prompt in Appendix~\ref{app:llm_prompt}. While encoder results in this paper are averaged over three seeds for scientific rigour, the submitted ensemble uses five seeds with per-instance majority voting.

Our best-performing system on the hidden set is GPT-5.2, achieving evasion-based clarity F1 = 0.74 (Task~1, ranked 22nd out of 44) and evasion F1 = 0.50 (Task~2, ranked 13th out of 33) \citep{thomas2026semeval2026task6clarity}. The RoBERTa-large ensemble achieved lower scores, with F1 = 0.72 for evasion-based clarity and F1 = 0.45 for evasion.\footnote{While only the best system was recorded on the leaderboard, our RoBERTa-large ensemble would have ranked approximately 25th out of 44 for Task~1 and 18th out of 33 for Task~2 if it had been submitted instead of our best system \citep{thomas2026semeval2026task6clarity}.} This reverses the trend observed on the public test set (Section~\ref{sec:results-encoder}), where encoder-based models outperform GPT-5.2.

Notably, both systems improved substantially from public to hidden test (RoBERTa-large clarity: 0.661 $\rightarrow$ 0.72; GPT-5.2 clarity: 0.626 $\rightarrow$ 0.74). Given the tight variance reported in Table~\ref{tab:encoder_results}, we do not attribute the encoder improvement only to the five-seed ensemble. We hypothesise that this gap may reflect distributional differences between the public and hidden test sets, potentially favouring labels on which models performed better. Additionally, GPT-5.2 outperforming the encoders on the hidden set suggests that the fine-tuned models may have partially overfitted to the training distribution, though the marked improvement of encoders on the hidden set complicates this interpretation.

%
%

\subsection{Error Analysis}
\label{subsec:results-error-analysis}

Figure~\ref{fig:confusion_matrix} shows row-normalised confusion matrices for both tasks for the  RoBERTa-large model, with per-label breakdowns in Appendix~\ref{app:per-label-analysis}.

At the clarity level, the \textit{Ambivalent Reply} class achieves the highest recall (0.79), followed by \textit{Clear Reply} (0.61) and \textit{Clear Non-Reply} (0.52). Most errors occur at the boundaries between \textit{Ambivalent Reply} and the other two classes, consistent with the \ac{iaa} patterns reported in Section~\ref{sec:class_imbalance}. At the evasion level, the model shows strong recognition for \textit{Clarification} (0.83) and \textit{Explicit} (0.60), but exhibits substantial confusion among \textit{Implicit}, \textit{General}, \textit{Deflection}, and \textit{Partial/half-answer}. This mirrors known areas of annotator disagreement and likely reflects semantic overlap between fine-grained evasion strategies, suggesting that further gains may require modelling annotation uncertainty (rather than only additional model optimisation).


%
%

\section{Conclusion}

In this paper, we presented our contribution to the CLARITY shared task at SemEval 2026. We compared fine-tuned encoder models with zero-shot decoder-only models. Evasion-based clarity performed comparably to direct clarity. RoBERTa-large was strongest on the public test set, whereas zero-shot GPT-5.2 generalised better on the hidden evaluation set, suggesting a trade-off between in-domain performance and robustness. Most auxiliary training variations we attempted yielded no consistent improvements.

Overall, detecting political evasion remains challenging, as reflected by annotator disagreement (Section~\ref{sec:class_imbalance}). Moreover, results on the test sets should be interpreted cautiously: several labels have low support, and macro F1 can be sensitive to a small number of predictions for rare classes (Table~\ref{tab:clarity_from_evasion_per_label}~and~\ref{tab:evasion_f1_per_label_support} in Appendix \ref{app:per-label-analysis}).

%
%

\section*{Limitations}
\label{sec:limitations}

\paragraph{Conflation in input representation analysis.}
The ablation discussed in Section~\ref{subsec:input-representation} and Appendix~\ref{app:input_rep} compares \textit{Segmented} and \textit{Marked} input formats. However, this comparison simultaneously varies two factors: the ordering of the question–answer fields and the mechanism used to mark their boundary. As a result, the observed performance differences cannot be attributed unambiguously to either factor. A more controlled analysis that isolates field ordering from boundary representation would provide clearer insight into which component drives the improvement. We do not pursue this analysis here, as it falls outside the primary scope of the shared task and our focus in this work.

\paragraph{Annotation and supervision signals.}
A further limitation concerns the supervision signals used for both tasks. For clarity classification, training relies on a single aggregated label per instance. Given the subjective nature of the task, soft labels that reflect the distribution of annotator judgements could better capture annotation uncertainty and potentially provide richer supervision. For the evasion task, multi-annotator labels are available only at test time, whereas training uses a single label per instance. This mismatch may underestimate model performance and fails to reflect the overlapping nature of evasion strategies, where multiple interpretations may be plausible. A multi-label formulation or training with annotator distributions could therefore provide a more faithful modelling framework.

\paragraph{Zero-shot evaluation of decoder-only models.}
In this work, decoder-only models were evaluated only in a zero-shot setting, without any supervised fine-tuning on the QEvasion dataset. While this design allows us to assess the out-of-the-box capabilities and cross-domain robustness of \acp{llm}, it does not reflect their full potential under task-specific training. In particular, parameter-efficient fine-tuning methods may enable these models to better capture the task taxonomy and domain-specific discourse patterns. Future work could therefore explore supervised fine-tuning of large decoder-only models, including parameter-efficient approaches such as LoRA \citep{hu2022lora}, as well as cross-domain transfer learning to improve the generalisability of encoder-based models.

\paragraph{Performance gap to top systems.}
The performance gap between our systems and the top-ranked submissions highlights a methodological limitation of the approaches we explored. In particular, our strongest systems rely on either single-pass encoder-based classification or zero-shot prompting, whereas the task overview \citep{thomas2026semeval2026task6clarity} indicates that the best performing submissions typically benefited from hierarchical decomposition, multi-stage inference, confidence-based routing, and more elaborate prompt design. These differences are especially relevant for evasion-level classification, where several categories are semantically close and were also reported by the organisers to be difficult both for systems and annotators to distinguish. Our models use the task taxonomy during training and label mapping, but they do not explicitly exploit that hierarchy at inference time through staged decision-making or branch-restricted prediction. As a result, they are likely less effective at resolving fine-grained distinctions such as those between \textit{General}, \textit{Deflection}, and \textit{Implicit}. However, higher benchmark performance from more complex pipelines does not necessarily indicate a more general solution: methods that are tightly coupled to the shared task taxonomy may achieve stronger benchmark performance while being less informative about transfer to other datasets or political settings. For this reason, we prioritised comparatively simple and reproducible methods, which we view as useful baselines even if they do not reach the performance of the most competitive task-specific systems.

\section*{Acknowledgments}

This work was supported by the Engineering and Physical Sciences Research Council [grant number EP/W524475/1]. The authors acknowledge the use of the Computational Research, Engineering and Technology Environment (CREATE) at King’s College London \cite{kings_create_2025}. The authors also thank the anonymous reviewers for their valuable comments and suggestions, which helped improve the quality of this work.

\bibliography{custom}

\clearpage

\appendix

\FloatBarrier

\section*{Appendix}

\section{QEvasion Dataset Analysis}
\label{apx:data-analysis}

As outlined in Section \ref{sec:background}, the CLARITY task comprises two hierarchically related classification tasks: (Task 1) clarity level classification and (Task 2) evasion level classification. An illustrative example of a question–answer pair with its associated labels is presented in Figure~\ref{fig:clarity-example}. The definitions of the class labels for both levels are presented in Table \ref{tab:label-definitions}.

\begin{figure}[t]
\centering

\begin{tcolorbox}[
  qabox,
  colback=questionbg,
  colframe=questionframe,
  title={\textcolor{questionframe}{\faMicrophone~\textbf{Interviewer's Question}}},
]
\small\sffamily
\textit{Can you share what you asked him about Afghanistan? What was your particular request for Afghanistan and the U.S. troops?}
\end{tcolorbox}


\begin{tcolorbox}[
  qabox,
  colback=answerbg,
  colframe=answerframe,
  title={\textcolor{answerframe}{\faComments~\textbf{President's Answer}}},
]
\small\sffamily
\textit{No, he asked us about Afghanistan. He said that he hopes that we're able to maintain some peace and security, and I said, That has a lot to do with you. He indicated that he was prepared to, quote, help on Afghanistan—I won't go into detail now; and help on Iran; and help on—and, in return, we told him what we wanted to do relative to bringing some stability and economic security or physical security to the people of Syria and Libya. So we had those discussions.}
\end{tcolorbox}


\begin{tcolorbox}[
  qabox,
  colback=labelbg,
  colframe=labelframe,
  title={\textcolor{labelframe}{\faTags~\textbf{Annotations}}},
]
\small
\badge{badgeclarity}{Clarity Level}{Ambivalent} \quad
\badge{badgeevasion}{Evasion Level}{Partial/half-answer} \quad
\badge{badgepresident}{President}{Biden} \quad
\badge{badgetrue}{Multiple Q.}{True} \quad
\badge{badgetrue}{Affirmative Q.}{True}
\end{tcolorbox}

\caption{An example from the CLARITY dataset}
\label{fig:clarity-example}
\end{figure}

\begin{figure}[!t]
    \centering
    \includegraphics[width=0.99\linewidth]{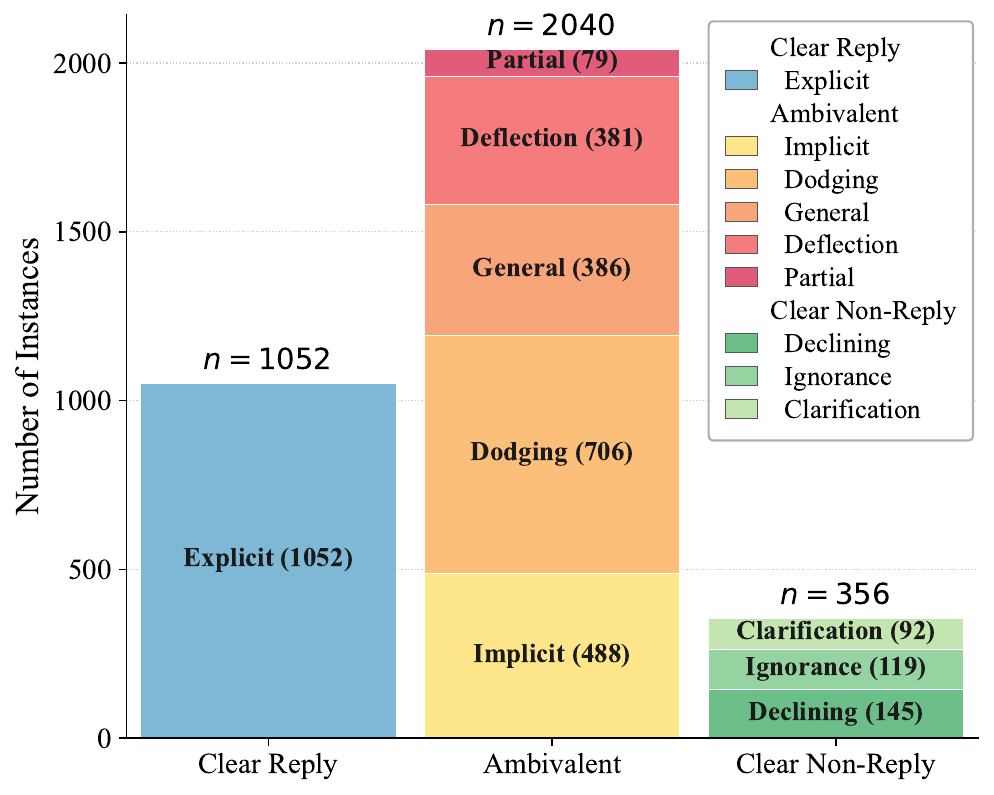}
    \caption{Class distribution in the training set.}
    \label{fig:overall_dististribution}
\end{figure}

Figure~\ref{fig:overall_dististribution} presents the distribution of clarity and evasion level classes in the training set. The most frequent clarity label is \textit{Ambivalent Reply} (59.2\%). Within this category, \textit{Dodging} is the most common evasion technique, whereas \textit{Partial/half-answer} is the least frequent. The second most prevalent clarity label is \textit{Clear Reply} (30.5\%), while \textit{Clear Non-Reply} (10.3\%) occurs least often. Overall, the figure indicates a substantial class imbalance in both tasks. 

A similar, although slightly different, distribution is observed in the public test set, where \textit{Ambivalent Reply} accounts for 66.9\% (206 instances), followed by \textit{Clear Reply} with 25.6\% (79 instances), and \textit{Clear Non-Reply} with 7.5\% (23 instances), confirming the persistence of class imbalance across data splits. 

For the evasion level, the test set includes three labels per instance, each provided by a different annotator. In 275 out of 308 cases (89.3\%), a final evasion label can be assigned through majority voting. However, in 33 instances (10.7\%), no majority label can be determined, as all three annotators assigned different labels. While our evaluation procedure preserves disagreement (Section~\ref{sec:eval_metrics}), the majority-vote breakdown above is useful for indicating how often a single-label collapse would require a tie-break.

\begin{figure}[!t]
    \centering
    \includegraphics[width=0.99\linewidth]{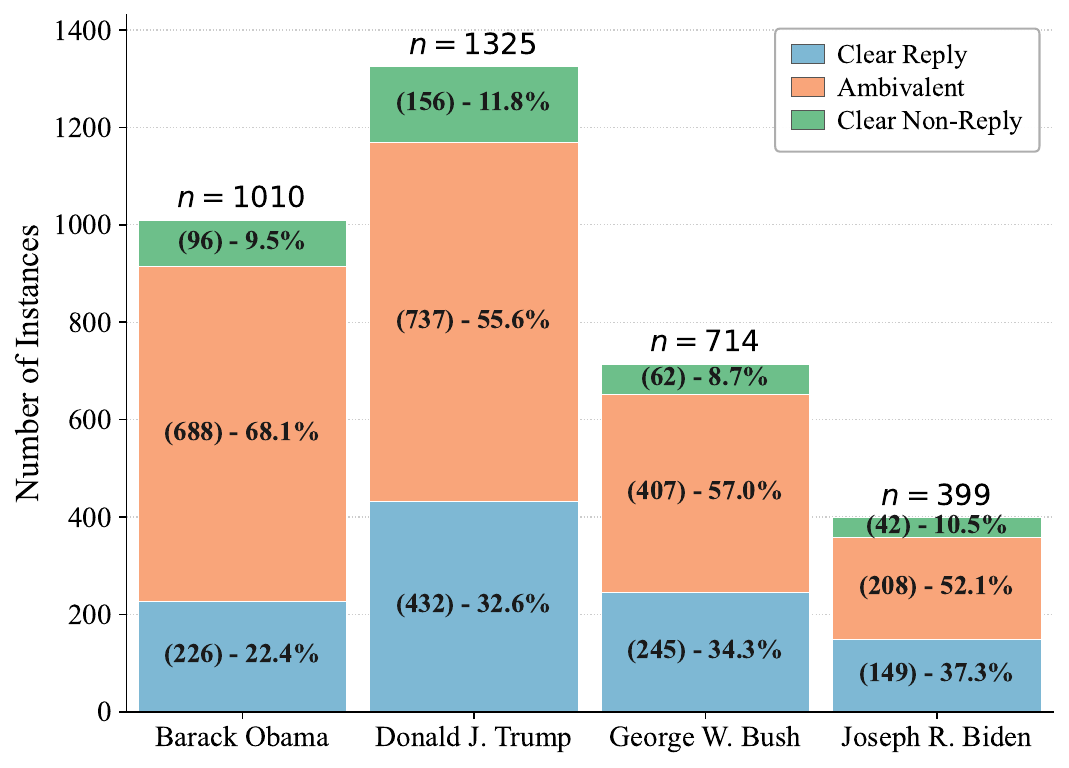}
    \caption{President and clarity level distribution in the training set.}
    \label{fig:president_dististribution}
\end{figure}

Figure~\ref{fig:president_dististribution} shows the distribution of presidents alongside the clarity level distribution in the training set. The most represented president is \textit{Trump} with 1{,}325 samples (38.4\%), followed by \textit{Obama} with 1{,}010 (29.3\%), \textit{Bush} with 714 (20.7\%), and \textit{Biden} with 399 (11.6\%). This indicates that the dataset is also imbalanced with respect to presidential interviews.

Overall, the four presidents exhibit similar proportions of \textit{Clear Reply}, \textit{Ambivalent Reply}, and \textit{Clear Non-Reply} instances. An exception is \textit{Obama}, who shows a higher percentage of ambivalent replies and a comparatively lower percentage of clear replies. President labels are not provided for the test set.

\begin{table*}[t]
\centering
\small
\setlength{\tabcolsep}{4pt} 
\renewcommand{\arraystretch}{1.05}

\begin{tabularx}{\textwidth}{@{}l >{\RaggedRight\arraybackslash}X @{}}
\toprule
\textbf{Label} & \textbf{Definition} \\
\midrule
\multicolumn{2}{@{}l}{\textit{Clarity Level}} \\
\midrule
\textbf{Clear Reply}      & Answers that admit a single interpretation. \\
\textbf{Clear Non-Reply}  & Answers that openly refuse to share information. \\
\textbf{Ambivalent Reply} & Valid answers that allow for multiple interpretations. \\
\midrule
\multicolumn{2}{@{}l}{\textit{Evasion Level}} \\
\midrule
\textbf{Explicit}      & The requested information is clearly and directly stated in the expected form. \\
\textbf{Implicit}      & The requested information is provided, but not explicitly stated or not presented in the expected form. \\
\textbf{Dodging}       & The question is completely ignored. \\
\textbf{General}       & A response is given, but it is overly general and lacks the specific details requested. \\
\textbf{Deflection}    & The response initially addresses the topic but then shifts focus, making a different point than the one asked. \\
\textbf{Partial}       & The response provides only part of the required information, addressing only a specific part of the request. \\
\textbf{Declining}     & The question is acknowledged, but the respondent directly or indirectly refuses to answer. \\
\textbf{Ignorance}     & The respondent explicitly states that they do not know the answer. \\
\textbf{Clarification} & The respondent does not provide the requested information and instead asks for clarification. \\
\bottomrule
\end{tabularx}

\caption{Clarity and evasion technique label definitions, adapted from \citet{thomas2024isaidthatdataset}.}
\label{tab:label-definitions}
\end{table*}

%
%

\section{Masking Ablation Study}
\label{sec:masking_appendix}

In the QEvasion dataset paper, the authors explored a `prior knowledge hypothesis', observing that clarity level classification performance was consistently higher on instances that did \textit{not} contain named entities \citep{thomas2024isaidthatdataset}. They attribute this effect to the fact that named entities often carry implicit, commonly assumed properties that are not explicitly stated in the response, therefore requiring models to rely on encoded world knowledge rather than surface-level discourse cues alone. As this effect was reported to be most pronounced for smaller models, we conducted a series of experiments using RoBERTa-base, both to reduce computational cost and to assess whether this phenomenon would persist in a more controlled setting. This observation motivated our decision to experiment with different masking strategies: (i) no masking, (ii) naive masking, and (iii) entity-aware masking. All masking experiments were conducted at the clarity level.\footnote{Masking implementation details are provided in the accompanying code.}

\subsection{Variants}

\paragraph{Naive masking.}

In the naive masking setting, person-name mentions are replaced with a single generic token \texttt{[PERSON]}. Person entities are identified using spaCy's named entity recogniser with the \texttt{en\_core\_web\_lg} model, and each detected mention is masked independently, without preserving identity across mentions or between the question and the answer. An example of naive masking is shown below:

\begin{list}{}{\leftmargin=1em \rightmargin=0em}
\raggedright
\item \textbf{Question:} Did John Smith support the proposal?
\item \textbf{Masked Question:} Did \texttt{[PERSON]} support the proposal?

\item \textbf{Answer:} John said that Mary Johnson was responsible.
\item \textbf{Masked Answer:} \texttt{[PERSON]} said that \texttt{[PERSON]} was responsible.
\end{list}

As all person mentions are mapped to the same placeholder, this approach removes explicit name information but collapses identity distinctions.

\paragraph{Entity-aware masking.}

Entity-aware masking extends naive masking by preserving consistency across references to the same individual within a question-answer pair. Person entities are again detected using spaCy's \texttt{en\_core\_web\_lg} model, but distinct placeholders of the form \texttt{[PERSON\_i]} are assigned to different individuals based on surface-form similarity. The same example under entity-aware masking becomes:

\begin{list}{}{\leftmargin=1em \rightmargin=0em}
\raggedright
\item \textbf{Question:} Did John Smith support the proposal?
\item \textbf{Masked Question:} Did \texttt{[PERSON\_1]} support the proposal?

\item \textbf{Answer:} John said that Mary Johnson was responsible.
\item \textbf{Masked Answer:} \texttt{[PERSON\_1]} said that \texttt{[PERSON\_2]} was responsible.
\end{list}

The underlying idea was to encourage the model to rely on relative reference patterns between the question and the answer, rather than on memorised world-level knowledge associated with specific entities.

\subsection{Results}

\begin{table}[t]
    \caption{\textbf{RoBERTa-base performance under different masking strategies for direct clarity classification on the public test split.}
    We report macro-F1 (mean $\pm$ standard deviation over three seeds).
    Experiments use a different training setup from the main baselines, so scores are not directly comparable across tables; comparisons within this table remain valid.}
  \centering
  \small
  \setlength{\tabcolsep}{6pt}
  \begin{threeparttable}
    \adjustbox{max width=\columnwidth}{
      \begin{tabular}{@{}lc@{}}
        \toprule
        \textbf{Masking Strategy} & \textbf{Macro-F1} \\
        \midrule
        None         & $0.585 \pm 0.013$ \\
        Naive        & $0.570 \pm 0.006$ \\
        Entity-aware & $0.566 \pm 0.016$ \\
        \bottomrule
      \end{tabular}
    }
  \end{threeparttable}
  \label{tab:masking_macro_f1_results}
\end{table}

Table~\ref{tab:masking_macro_f1_results} reports RoBERTa-base performance under the three masking settings, with all other training choices held constant and only the masking strategy varied. Across runs, the unmasked baseline attains the highest mean macro-F1 ($0.585 \pm 0.013$), while both masking variants yield slightly lower mean scores on the HuggingFace test split (naive: $0.570 \pm 0.006$, entity-aware: $0.566 \pm 0.016$). Overall, these results suggest that, for this encoder model and training setup, masking person entities does not provide a clear benefit for clarity-level classification. Differences between the two masking variants are modest, and we do not observe a consistent advantage of entity-aware over naive masking in this setting.

However, these results should not be taken as direct evidence against the task paper’s `prior-knowledge hypothesis'. Our masking only replaces \textit{person} names, and it depends on automatic entity detection and fairly simple matching, which can introduce noise and remove useful cues (for example, who is being referred to across the question and answer - especially for naive masking). Also, lower macro-F1 on this test split does not rule out the possibility that masking could help in other settings - for instance, when evaluating on interviews from different sources or speakers - which we leave for future work.

%
%

\section{Loss Weighting Ablation Study}
\label{app:loss_weighting}

As introduced in Section~\ref{sec:loss_weighting}, we evaluate RoBERTa-large under three loss-weighting strategies: (i) \textit{Unweighted}, (ii) \textit{Balanced}, and (iii) \textit{Sqrt}, to account for class imbalance.

Table~\ref{tab:roberta_large_weighting_ablation} reports results on the public test set. Overall, weighting does not yield consistent improvements. The unweighted model performs best, achieving an evasion-based clarity macro F1 of 0.661. For evasion classification, it attains $\mathrm{ACC}_{\mathrm{match}} = 0.539$ and an average macro F1 of 0.371 across annotators. Both weighting schemes reduce performance on average and increase variance, particularly under the more aggressive balanced setting.

\begin{table*}[t]
  \centering

\caption{\textbf{RoBERTa-large loss-weighting ablation on the public test set.}
For evasion-based clarity we report macro-F1 (F1), precision (P), and recall (R).
For evasion we report $\mathrm{ACC}_{\mathrm{match}}$, the fraction of predictions matching at least one annotator,
per-annotator macro-F1, and the average macro-F1 across annotators.
All models are fine-tuned on evasion labels, with clarity inferred via the taxonomy.
Results are averaged over three seeds (mean on the main row, standard deviation in brackets on the row beneath).
Best-performing metrics are shown in bold.}
  \label{tab:roberta_large_weighting_ablation}

  \small
  \begin{tabular*}{\textwidth}{@{\extracolsep{\fill}} l ccc c cccc}
    \toprule
    \multirow{2}{*}{\textbf{Loss}} & \multicolumn{3}{c}{\textit{Evasion-based clarity}} & \multicolumn{5}{c}{\textit{Evasion}} \\
    \cmidrule(lr){2-4} \cmidrule(lr){5-9}
    & \textbf{F1} & \textbf{P} & \textbf{R} & $\mathbf{ACC}_{\mathbf{match}}$ & \textbf{F1\textsubscript{A1}} & \textbf{F1\textsubscript{A2}} & \textbf{F1\textsubscript{A3}} & \textbf{F1\textsubscript{avg}} \\
    \midrule

    Unweighted
      & \textbf{0.661} & \textbf{0.694} & 0.641 & \textbf{0.539} & \textbf{0.363} & \textbf{0.378} & \textbf{0.374} & \textbf{0.371} \\[-0.2ex]
      & \sd{0.022} & \sd{0.013} & \sd{0.023} & \sd{0.011} & \sd{0.030} & \sd{0.035} & \sd{0.047} & \sd{0.037} \\[0.65ex]

    Balanced
      & 0.603 & 0.602 & \textbf{0.642} & 0.515 & 0.350 & 0.347 & 0.327 & 0.341 \\[-0.2ex]
      & \sd{0.077} & \sd{0.098} & \sd{0.023} & \sd{0.049} & \sd{0.033} & \sd{0.035} & \sd{0.047} & \sd{0.038} \\[0.65ex]

    Sqrt
      & 0.602 & 0.617 & 0.625 & 0.505 & 0.318 & 0.304 & 0.302 & 0.308 \\[-0.2ex]
      & \sd{0.046} & \sd{0.052} & \sd{0.017} & \sd{0.024} & \sd{0.036} & \sd{0.041} & \sd{0.034} & \sd{0.035} \\

    \bottomrule
  \end{tabular*}
\end{table*}

%
%

\section{Input Representation Ablation Study}
\label{app:input_rep}

Table~\ref{tab:roberta_base_input_ablation} compares the two input formats described in Section~\ref{subsec:input-representation} for RoBERTa-base. The marked representation (\texttt{[QUESTION]} q \texttt{[ANSWER]} a) consistently outperforms the segmented format (\texttt{[CLS]} a \texttt{[SEP]} q \texttt{[SEP]}) on both tasks, improving evasion-based clarity macro F1 from 0.518 to 0.595 and evasion F1\textsubscript{avg} from 0.270 to 0.338. The same trend holds for direct clarity prediction (Table~\ref{tab:roberta_base_direct_clarity_ablation}), where the marked format improves macro F1 from 0.573 to 0.598. This suggests that explicit boundary tokens and/or presenting the question before the answer provides a stronger inductive bias for modelling question-answer alignment. As the creators of the QEvasion dataset note, limited context windows may cap encoder performance (512 tokens in this case for RoBERTa-base) \citep{thomas2024isaidthatdataset}. One possible explanation is that placing the question before the answer reduces the likelihood that the question - the primary grounding signal - is truncated when inputs exceed the maximum length, even if this increases truncation of the answer. We leave a more controlled analysis of truncation effects (e.g., varying maximum length and truncation strategy) to future work.

\begin{table*}[t]
  \centering
  \caption{\textbf{RoBERTa-base input representation ablation on the public test set.}
For evasion-based clarity we report macro-F1 (F1), precision (P), and recall (R).
For evasion we report $\mathrm{ACC}_{\mathrm{match}}$, the fraction of predictions matching at least one annotator,
per-annotator macro-F1, and the average macro-F1 across annotators.
All models are fine-tuned on evasion labels, with clarity inferred via the taxonomy.
Results are averaged over three seeds (mean on the main row, standard deviation in brackets on the row beneath).
Best-performing metrics are shown in bold.}
  \label{tab:roberta_base_input_ablation}

  \small
 \begin{tabular*}{\textwidth}{@{\extracolsep{\fill}} l ccc c cccc}
    \toprule
    \multirow{2}{*}{\textbf{Input}} & \multicolumn{3}{c}{\textit{Evasion-based clarity}} & \multicolumn{5}{c}{\textit{Evasion}} \\
    \cmidrule(lr){2-4} \cmidrule(lr){5-9}
    & \textbf{F1} & \textbf{P} & \textbf{R} & $\mathbf{ACC}_{\mathbf{match}}$ & \textbf{F1\textsubscript{A1}} & \textbf{F1\textsubscript{A2}} & \textbf{F1\textsubscript{A3}} & \textbf{F1\textsubscript{avg}} \\
    \midrule

    Segmented
      & 0.518 & 0.520 & 0.534 & 0.473 & 0.277 & 0.272 & 0.261 & 0.270 \\[-0.2ex]
      & \sd{0.019} & \sd{0.018} & \sd{0.021} & \sd{0.004} & \sd{0.015} & \sd{0.016} & \sd{0.003} & \sd{0.004} \\[0.65ex]

    Marked
      & \textbf{0.595} & \textbf{0.589} & \textbf{0.619} & \textbf{0.516} & \textbf{0.349} & \textbf{0.332} & \textbf{0.333} & \textbf{0.338} \\[-0.2ex]
      & \sd{0.028} & \sd{0.035} & \sd{0.030} & \sd{0.014} & \sd{0.024} & \sd{0.007} & \sd{0.008} & \sd{0.011} \\

    \bottomrule
  \end{tabular*}
\end{table*}

\begin{table}[t]
  \centering
  \caption{\textbf{RoBERTa-base input representation ablation for direct clarity on the public test set.}
We report macro-F1 (F1), precision (P), and recall (R).
Results are averaged over three seeds (mean on the main row, standard deviation in brackets on the row beneath).
Best-performing metrics are shown in bold.}
  \label{tab:roberta_base_direct_clarity_ablation}

  \small
  \begin{tabular*}{\columnwidth}{@{\extracolsep{\fill}} l ccc}
    \toprule
    \multirow{2}{*}{\textbf{Input}} & \multicolumn{3}{c}{\textit{Direct clarity}} \\
    \cmidrule(lr){2-4}
    & \textbf{F1} & \textbf{P} & \textbf{R} \\
    \midrule

    Segmented
      & 0.573 & 0.552 & 0.621 \\[-0.2ex]
      & \sd{0.010} & \sd{0.018} & \sd{0.020} \\[0.65ex]

    Marked
      & \textbf{0.598} & \textbf{0.586} & \textbf{0.636} \\[-0.2ex]
      & \sd{0.010} & \sd{0.013} & \sd{0.021} \\

    \bottomrule
  \end{tabular*}
\end{table}

%
%

\section{Stratified vs President-based Splitting Ablation Study}
\label{app:strat_vs_president_disjoint}

As described in Section~\ref{sec:data_splits}, we evaluate both label-stratified and president-disjoint splitting strategies. In the label-stratified setting, we apply \textit{dual stratification} over both clarity and evasion labels. In the president-disjoint setting, all responses from a given president appear exclusively in one split, preventing speaker leakage.

Table~\ref{tab:roberta_large_split_ablation} reports RoBERTa-large performance under the two strategies. The stratified split outperforms the president-disjoint split by approximately 0.04 in evasion-based clarity F1 and 0.05 in evasion F1\textsubscript{avg}. This gap likely reflects the slight variation in label distributions across presidents noted in Appendix~\ref{apx:data-analysis}. We note that the president-disjoint constraint applies only to the train/validation partition; the test set is shared across both configurations. The observed performance difference therefore primarily reflects the effect of a stricter model selection criterion rather than a direct measure of cross-speaker generalisation. A fully disjoint evaluation pipeline, including the test set, is left to future work.

\begin{table*}[t]
  \centering
  
  \caption{\textbf{RoBERTa-large split-strategy ablation on the public test set.}
For evasion-based clarity we report macro-F1 (F1), precision (P), and recall (R).
For evasion we report $\mathrm{ACC}_{\mathrm{match}}$, the fraction of predictions matching at least one annotator,
per-annotator macro-F1, and the average macro-F1 across annotators.
All models are fine-tuned on evasion labels, with clarity inferred via the taxonomy.
Results are averaged over three seeds (mean on the main row, standard deviation in brackets on the row beneath).
Best-performing metrics are shown in bold.}
  
  \label{tab:roberta_large_split_ablation}
  \small
  \begin{tabular*}{\textwidth}{@{\extracolsep{\fill}} l ccc c cccc}
    \toprule
    \multirow{2}{*}{\textbf{Split Method}} & \multicolumn{3}{c}{\textit{Evasion-based clarity}} & \multicolumn{5}{c}{\textit{Evasion}} \\
    \cmidrule(lr){2-4} \cmidrule(lr){5-9}
    & \textbf{F1} & \textbf{P} & \textbf{R} & $\mathbf{ACC}_{\mathbf{match}}$ & \textbf{F1\textsubscript{A1}} & \textbf{F1\textsubscript{A2}} & \textbf{F1\textsubscript{A3}} & \textbf{F1\textsubscript{avg}} \\
    \midrule
    Label-stratified
      & \textbf{0.661} & \textbf{0.694} & \textbf{0.641} & \textbf{0.539} & \textbf{0.363} & \textbf{0.378} & \textbf{0.374} & \textbf{0.371} \\[-0.2ex]
      & \sd{0.022} & \sd{0.013} & \sd{0.023} & \sd{0.011} & \sd{0.030} & \sd{0.035} & \sd{0.047} & \sd{0.037} \\[0.65ex]
    President disjoint
      & 0.624 & 0.642 & 0.624 & 0.529 & 0.330 & 0.306 & 0.318 & 0.318 \\[-0.2ex]
      & \sd{0.029} & \sd{0.027} & \sd{0.035} & \sd{0.012} & \sd{0.036} & \sd{0.033} & \sd{0.058} & \sd{0.039} \\
    \bottomrule
  \end{tabular*}
\end{table*}

%
%

\section{Per-label Performance Analysis}
\label{app:per-label-analysis}

Table~\ref{tab:clarity_from_evasion_per_label} reports precision, recall, and F1 per clarity label for the best-performing RoBERTa-large model. The \textit{Ambivalent} class achieves the strongest results (F1 = 0.798), consistent with its larger support (206 instances). The \textit{Clear Non-Reply} class shows high precision but low recall (0.522), suggesting the model tends to miss instances of this class, likely due to its limited support (23 instances). Finally, the \textit{Clear Reply} class presents the weakest overall F1 (0.577), with relatively balanced but moderate precision and recall.

Table~\ref{tab:evasion_f1_per_label_support} reports the per-label support and F1 scores for each annotator separately and their average for the best-performing RoBERTa-large model. The \textit{Clarification} class achieves the highest average F1 (0.841). However, its very low support (4 instances per annotator) limits the reliability of this estimate. \textit{Explicit} is the strongest well-supported class (0.566), followed by \textit{Declining to answer} (0.424) and \textit{Claims ignorance} (0.447). The most problematic classes are \textit{Deflection} (0.084), \textit{General} (0.184), and \textit{Implicit} (0.277), all of which suffer from both low F1 and high variance across annotators, reflecting their inherent ambiguity and annotation inconsistency. All these classes belong to the ambivalent clarity level. Finally, \textit{Partial/half-answer} is also poorly recognised (0.113), likely due to its extremely limited support across all annotators.

\begin{table*}[t]
  \centering
  \caption{\textbf{Per-label clarity (evasion-based) on the test set (RoBERTa-large).}
  Clarity labels are inferred from evasion via the taxonomy. Support is the number of test instances per label.
  Results are averaged over 3 seeds (13, 21, 42): mean on the main row, sample standard deviation in brackets on the row beneath.}
  \label{tab:clarity_from_evasion_per_label}

  \small
  \setlength{\tabcolsep}{10pt}
  \begin{tabular*}{\textwidth}{@{\extracolsep{\fill}} l c c c c}
    \toprule
    \textbf{Clarity label} & \textbf{Support} & \textbf{P} & \textbf{R} & \textbf{F1} \\
    \midrule

    Clear Reply
      & 79  & 0.550 & 0.608 & 0.577 \\[-0.2ex]
      &     & \sd{0.005} & \sd{0.046} & \sd{0.019} \\[0.65ex]

    Ambivalent
      & 206 & 0.801 & 0.794 & 0.798 \\[-0.2ex]
      &     & \sd{0.003} & \sd{0.022} & \sd{0.011} \\[0.65ex]

    Clear Non-Reply
      & 23  & 0.733 & 0.522 & 0.608 \\[-0.2ex]
      &     & \sd{0.039} & \sd{0.087} & \sd{0.072} \\[0.65ex]

    \bottomrule
  \end{tabular*}
\end{table*}

\begin{table*}[t]
  \centering
  \caption{\textbf{Per-label evasion F1 on the test set (RoBERTa-large).}
  For each evasion label, we report F1 against each annotator (A1-A3) and the average across annotators.
  Support indicates the number of test instances for that label for each annotator.
  Results are averaged over 3 seeds (13, 21, 42): mean on the main row, sample standard deviation in brackets on the row beneath.}
  \label{tab:evasion_f1_per_label_support}

  \small
  \setlength{\tabcolsep}{6pt}
  \begin{tabular*}{\textwidth}{@{\extracolsep{\fill}} l ccc cccc}
    \toprule
    \multirow{2}{*}{\textbf{Evasion label}}
      & \multicolumn{3}{c}{\textit{Support}}
      & \multicolumn{4}{c}{\textit{F1}} \\
    \cmidrule(lr){2-4} \cmidrule(lr){5-8}
      & \textbf{A1} & \textbf{A2} & \textbf{A3}
      & \textbf{F1\textsubscript{A1}} & \textbf{F1\textsubscript{A2}} & \textbf{F1\textsubscript{A3}} & \textbf{F1\textsubscript{avg}} \\
    \midrule

    Explicit
      & 106 & 53 & 80
      & 0.589 & 0.536 & 0.573 & 0.566 \\[-0.2ex]
      &     &    &
      & \sd{0.033} & \sd{0.027} & \sd{0.028} & \sd{0.030} \\[0.65ex]

    Implicit
      & 54 & 54 & 67
      & 0.242 & 0.257 & 0.333 & 0.277 \\[-0.2ex]
      &    &    &
      & \sd{0.083} & \sd{0.048} & \sd{0.077} & \sd{0.069} \\[0.65ex]

    Dodging
      & 58 & 72 & 43
      & 0.469 & 0.371 & 0.381 & 0.407 \\[-0.2ex]
      &    &    &
      & \sd{0.015} & \sd{0.035} & \sd{0.029} & \sd{0.021} \\[0.65ex]

    General
      & 29 & 78 & 65
      & 0.174 & 0.153 & 0.224 & 0.184 \\[-0.2ex]
      &    &    &
      & \sd{0.019} & \sd{0.050} & \sd{0.055} & \sd{0.029} \\[0.65ex]

    Deflection
      & 30 & 22 & 23
      & 0.062 & 0.093 & 0.098 & 0.084 \\[-0.2ex]
      &    &    &
      & \sd{0.023} & \sd{0.017} & \sd{0.047} & \sd{0.018} \\[0.65ex]

    Partial/half-answer
      & 8 & 5 & 5
      & 0.095 & 0.122 & 0.122 & 0.113 \\[-0.2ex]
      &   &   &
      & \sd{0.086} & \sd{0.113} & \sd{0.113} & \sd{0.104} \\[0.65ex]

    Declining to answer
      & 10 & 9 & 14
      & 0.383 & 0.519 & 0.370 & 0.424 \\[-0.2ex]
      &    &   &
      & \sd{0.102} & \sd{0.129} & \sd{0.080} & \sd{0.096} \\[0.65ex]

    Claims ignorance
      & 9 & 11 & 7
      & 0.411 & 0.508 & 0.422 & 0.447 \\[-0.2ex]
      &   &    &
      & \sd{0.070} & \sd{0.226} & \sd{0.102} & \sd{0.122} \\[0.65ex]

    Clarification
      & 4 & 4 & 4
      & 0.841 & 0.841 & 0.841 & 0.841 \\[-0.2ex]
      &   &   &
      & \sd{0.167} & \sd{0.167} & \sd{0.167} & \sd{0.167} \\

    \bottomrule
  \end{tabular*}
\end{table*}

%
%

\section{Exploratory Experiments}
\label{sec:exploratory}

%
%

\subsection{Cross-Domain Transfer}
\label{app:exploratory_cross_domain}

As performance gains from training exclusively on the QEvasion dataset began to plateau during development, we explored the use of an additional dataset from a different domain, Earnings Calls Q\&A\footnote{Earnings Calls Q\&A consists of question-answer exchanges between financial analysts and company executives during quarterly earnings announcements.} \citep{nuaimi-etal-2025-detecting}. This dataset employs a Rasiah-style taxonomy \citep{RASIAH2010664}, together with Bavelas forms \citep{bavelas1990equivocal} and Bull subtypes \citep{bull1998equivocation}, maintaining continuity with established frameworks for analysing psychological and political equivocation. Importantly, its labelling scheme is derived from the same theoretical foundations as QEvasion, making Earnings Calls Q\&A a strong candidate as a complementary resource. In particular, the Rasiah-style taxonomy uses the labels \textit{Direct}, \textit{Intermediate}, \textit{Fully Evasive}, which align closely with QEvasion's clarity-level labels \textit{Clear Reply}, \textit{Ambivalent Reply}, and \textit{Clear Non-Reply}, respectively.

Although the label sets are conceptually aligned, preliminary intermediate-task fine-tuning experiments were sensitive to optimisation settings and consistently reduced QEvasion validation macro-F1 in our setup (RoBERTa-large; Earnings Calls Q\&A with Rasiah labels: 5 epochs; learning rate $3\times10^{-5}$; batch size 32; bf16; followed by continued fine-tuning on QEvasion). We did not investigate further whether this reflects domain mismatch or suboptimal training settings, and therefore excluded these results from the final system.

%
%

\subsection{Cognitive Distortion Presence as an Auxiliary Signal}
\label{app:cd}

\Acp{cd} are systematic patterns of biased reasoning studied in \ac{cbt}. A growing body of work in \ac{nlp} aims to detect them automatically in text \citep{sage-etal-2025-survey}. We hypothesised that distorted reasoning patterns in political responses - such as \textit{Overgeneralisation} or \textit{All-or-Nothing Thinking} - might provide a useful auxiliary signal for evasion detection, since evasive answers may exhibit superficially similar rhetorical patterns (e.g., vague generalisations or deflective framing).

To test this, we trained a lightweight \ac{cd} detector using the \ac{cd} dataset TherapistQA \citep{shreevastava_2021_detecting}. We encoded texts with Sentence-BERT and fit a logistic regression model to predict binary distortion presence. We then scored each QEvasion training instance and discretised the resulting probabilities into two buckets (\texttt{CD\_LOW} and \texttt{CD\_HIGH}) using a threshold derived from the training distribution. These bucket tokens were prepended to the encoder input, allowing the model to condition on the estimated distortion level with minimal architectural changes.

In practice, this signal provided little discriminative value for either clarity or evasion classification. This may reflect a substantial domain gap between therapeutic dialogue and political interviews, where linguistic cues for \acp{cd} may not transfer cleanly. Additionally, the discretised token introduced noise and increased instability across seeds. Given these results, we excluded this approach from our final systems. It is also important to note that discretising a scalar probability into two buckets is a blunt instrument and likely discards useful information (e.g., a continuous score or predicted \ac{cd} type); we leave a more detailed investigation to future work.

%
%

\section{Hyperparameters and Training Setup}
\label{sec:training_details_appendix}

In Table~\ref{tab:encoder_training_details}, we report the hyperparameters and optimisation settings used across all encoder experiments. To ensure comparability, we fixed a single set of hyperparameters based on standard practice and applied them uniformly across all models, without any model-specific tuning. Full implementation details are available in our released codebase.

\begin{table*}[t]
\centering
\small
\caption{\textbf{Training and optimisation settings for all encoder experiments.} All hyperparameters are shared across models except batch size and gradient accumulation steps, which are reduced for DeBERTa-v3-large to accommodate its larger memory footprint (effective batch size 32 throughout).}
\label{tab:encoder_training_details}
\setlength{\tabcolsep}{6pt}
\begin{tabular*}{\textwidth}{@{\extracolsep{\fill}}lcccc@{}}
\toprule
\textbf{Setting} &
\textbf{RoBERTa-base} &
\textbf{RoBERTa-large} &
\textbf{DeBERTa-v3-base} &
\textbf{DeBERTa-v3-large} \\
\midrule
Max input length      & \multicolumn{4}{c}{512} \\
Learning rate         & \multicolumn{4}{c}{$2 \times 10^{-5}$} \\
Warmup ratio          & \multicolumn{4}{c}{0.1} \\
Weight decay          & \multicolumn{4}{c}{0.01} \\
Dropout               & \multicolumn{4}{c}{0.1} \\
Max epochs            & \multicolumn{4}{c}{20} \\
Precision             & \multicolumn{4}{c}{bfloat16} \\
\midrule
Batch size (train / eval)  & 32 / 32 & 32 / 32 & 32 / 32 & 16 / 16 \\
Gradient accumulation      & 1       & 1       & 1       & 2       \\
\midrule
Checkpoint selection  & \multicolumn{4}{c}{Best macro F1 (validation)} \\
Early stopping        & \multicolumn{4}{c}{Patience 5, threshold $10^{-3}$} \\
Seeds                 & \multicolumn{4}{c}{(13, 21, 42)} \\
\bottomrule
\end{tabular*}

\end{table*}

%
%

\section{Random Seeds}
\label{sec:seeds_appendix}

All encoder experiments are conducted using three fixed random seeds $(13, 21, 42)$. Results are reported as the mean across seeds, with the standard deviation shown in brackets in the tables. The random seed controls sources of stochasticity during training.

%
%

\section{Code, Compute, and Reproducibility}
\label{sec:code_and_repro_appendix}

\paragraph{Code release and reproducibility.}
In line with recent calls for improved reproducibility in machine learning research \citep{semmelrock_2025_reproducibilitymachinelearningbasedresearch, pineau2020improvingreproducibilitymachinelearning}, we release the full codebase upon publication to support reproducibility.

\paragraph{Compute.}
All encoder-based models reported in this paper were trained on a single NVIDIA A100-SXM4-40GB GPU. The total cumulative training runtime across all encoder experiments and multi-seed runs reported in this paper was $22{,}049$ seconds (approximately $6.1$ hours). For zero-shot experiments, we queried decoder-only models via hosted inference APIs (including the OpenAI API\footnote{\url{https://developers.openai.com/api/docs/}} for GPT5.2 and the Hugging Face Inference API\footnote{\url{https://huggingface.co/docs/inference-providers/en/index}} for open-weight models) and did not perform any additional fine-tuning.

\section{LLM Prompt}
\label{app:llm_prompt}

Figure~\ref{fig:prompt_taskb} shows the system prompt used for zero-shot evasion classification. The prompt instructs the model to assign exactly one of the nine evasion labels to each input question-answer pair, following a structured decision ladder with tie-breaking rules and in-context mini-examples. To improve throughput, inference was performed in batches, with the model receiving multiple question-answer pairs in a single API call and returning predictions as a JSON array. The evasion-based clarity label was then derived deterministically from the predicted evasion label via the taxonomy. 

\begin{figure*}[t]
\centering
\begin{adjustbox}{width=\linewidth}
\begin{minipage}{\linewidth}
\begin{lstlisting}[style=aclprompt]
system_prompt: |
  You are an expert annotator for CLARITY Task B (evasion level).
  Input: a QUESTION and an ANSWER (political interview style).
  Output: EXACTLY ONE label for each item from this set:
  Explicit, Implicit, Dodging, General, Deflection,
  Partial/half-answer, Declining to answer, Claims ignorance, Clarification.

  Core principle: decide based on whether the ANSWER supplies the *requested commitment*.
  Requested commitment = the specific yes/no, person, time, place, number, policy stance,
  or concrete plan the QUESTION asks for.
  If that commitment is present (even indirectly), it is NOT Dodging/Deflection.

  Step 0 - normalise the question:
  - Treat multi-part questions as requiring ALL parts unless the question clearly
    foregrounds one part.
  - If the question contains a yes/no + 'why/how/what specifics', then a bare
    yes/no is incomplete.

  Decision ladder (apply in order; stop at the first that clearly applies):
  1) Clarification
     - The answer primarily asks the interviewer to repeat/clarify/restate.
     - If it both asks for clarification AND gives the requested commitment,
       choose based on the dominant function.

  2) Claims ignorance
     - The answer asserts lack of knowledge/recall/awareness.
     - IMPORTANT: If the answer later gives the requested commitment anyway,
       do NOT pick Claims ignorance.

  3) Declining to answer
     - The answer refuses to provide the requested commitment now.
     - If it refuses but then gives the requested commitment, label by what dominates.

  4) Explicit
     - Directly states the requested commitment (even if brief), in the form requested.

  5) Implicit
     - The requested commitment is not stated verbatim, but is clearly recoverable
       via a straightforward inference.
     - Test: a reasonable listener could paraphrase the commitment in one sentence
       without guessing.

  6) Partial/half-answer
     - Answers ONE required part but omits other required parts.

  7) General vs Deflection vs Dodging (non-answer family)
     7a) General    - On-topic, but avoids the requested commitment by staying
                      vague or non-committal.
     7b) Deflection - Acknowledges the question then shifts to a different frame.
                      No requested commitment.
     7c) Dodging    - Does NOT engage the requested commitment AND does NOT
                      meaningfully address the topic.

  Tie-breakers:
  - Concrete list of steps in response to 'what will you do'
    -> prefer Explicit (or Partial if incomplete).
  - Gives constraints but never the asked commitment -> Declining to answer.
  - Explains the topic but not the specific asked stance -> General or Deflection,
    not Dodging.
  - 'I already answered that' without the commitment -> Declining to answer.

  Mini-examples:
  A) Dodging vs Deflection
  Q: Did you meet the leader yesterday?
  A (Dodging):     Our country is doing very well economically.
  A (Deflection):  I understand the question - what matters is focusing on jobs
                   for families.

  B) General vs Implicit
  Q: Do you support the bill?
  A (General):  We're considering all options and working with colleagues.
  A (Implicit): I'll be voting yes when it reaches the floor.

  C) Declining vs General
  Q: What concessions do you want them to make?
  A (Declining): I'll tell you after the meeting.
  A (General):   We want cooperation and stability.

  D) Claims ignorance vs General
  Q: Have you seen evidence of acceleration?
  A (Claims ignorance): I haven't seen that report.
  A (General):          We're concerned and monitoring closely.

  Output rules:
  - Choose exactly one label per item.
  - Keep the item order exactly.
  - Judge using only the provided QUESTION and ANSWER.

  Respond ONLY with a JSON object in this exact format: {"labels": ["label1", "label2", ...]}
  No explanation, no markdown fences, only the JSON.
\end{lstlisting}
\end{minipage}
\end{adjustbox}
\caption{Zero-shot prompt for evasion label prediction and evasion-based clarity.}
\label{fig:prompt_taskb}
\end{figure*}

\end{document}